# Optimizing Data Augmentation for Real-Time Small UAV Detection: A Lightweight Context-Aware Approach

Amir Zamani
*Department of Computer Engineering*
*Comprehensive University of the Islamic Revolution*
Tehran, Iran
amirzamani@cuir.ac.ir

Zeinab Abedini
*Department of Computer Engineering*
*Sharif University of Technology*
Tehran, Iran
z.abedini@sharif.edu

***Abstract*— Visual detection of Unmanned Aerial Vehicles (UAVs) is a critical task in surveillance systems due to their small physical size and environmental challenges. Although deep learning models have achieved significant progress, deploying them on edge devices necessitates the use of lightweight models, such as YOLOv11 Nano, which possess limited learning capacity. In this research, an efficient and context-aware data augmentation pipeline, combining Mosaic strategies and HSV color-space adaptation, is proposed to enhance the performance of these models. Experimental results on four standard datasets demonstrate that the proposed approach, compared to heavy and instance-level methods like Copy-Paste, not only prevents the generation of synthetic artifacts and overfitting but also significantly improves mean Average Precision (mAP) across all scenarios. Furthermore, the evaluation of generalization capability under foggy conditions revealed that the proposed method offers the optimal balance between Precision and stability for real-time systems, whereas alternative methods, such as MixUp, are effective only in specific applications.**

## I. Introduction

In recent years, Unmanned Aerial Vehicles (UAVs), commonly known as drones, have played a pivotal role in a wide range of civil and military applications, including infrastructure monitoring, traffic surveillance, and search-and-rescue operations [1], [2]. However, the rapid expansion and easy access to this technology have raised serious concerns regarding security and privacy. Unauthorized flights in sensitive areas can pose a threat to public safety and critical infrastructure; therefore, developing real-time and accurate detection systems to identify and neutralize such threats is essential [3].

Visual detection of UAVs inherently faces multiple technical challenges. On the one hand, UAVs have small physical dimensions and, at long distances, occupy only a minimal portion of image pixels, leading to a loss of textural information [4]. On the other hand, their visual and behavioral similarity to birds increases the false alarm rate in surveillance systems [5]. Although deep learning-based models, particularly Convolutional Neural Networks (CNNs) [6], [7], have made significant progress in detection accuracy, most existing research focuses on high-capacity models that require powerful Graphics Processing Units (GPUs) [8]. In operational applications, anti-drone systems are often deployed on embedded platforms and Edge Devices such as NVIDIA Jetson or Raspberry Pi, which face severe constraints in processing power and energy consumption. Recent studies have shown that deploying heavy models on these platforms poses serious challenges in terms of Frame Rate (FPS) [9], [10]. These limitations highlight the need to utilize lightweight models such as YOLOv11 Nano [11], which are optimized for ultra-fast inference. However, because they have fewer parameters, lightweight models struggle to extract features from small objects and to separate them from complex backgrounds.

One effective solution to overcome the lack of training data and improve model performance is to use data augmentation techniques. Recent studies have emphasized the effectiveness of Copy-Paste strategies [12], [13]. In this method, object instances are cropped from source images and pasted into random positions on background images to increase data diversity. Although this approach is efficient for high-capacity models, empirical observations show that it is challenging for Nano models. Copy-Paste methods often introduce synthetic artifacts and discontinuities in image texture; lightweight models may overfit to these synthetic noises rather than learning the main features of the drone.

In this research, focusing on hardware limitations, an efficient and context-aware data augmentation optimization strategy is proposed. The proposed approach combines the Mosaic method [14] with HSV color space adaptation. Unlike Copy-Paste methods, this approach preserves image texture continuity and simulates diverse lighting conditions, significantly increasing the generalization capability of the YOLOv11 Nano model. The main innovation of this research is the critical evaluation of the impact of data augmentation methods on Nano models and the empirical proof of the inefficiency of instance-level methods for this specific architecture. For a comprehensive assessment, four valid datasets, including Drone-vs-Bird [5], DUT-Anti-UAV [8], Det-Fly [4], and the Foggy Drone Dataset (FDD) [13], have been used, covering a wide range of environmental challenges.

The remainder of this paper is organized as follows: Section II reviews related work. Section III details the proposed context-aware pipeline. Section IV presents the experimental setup and quantitative results. Section V provides a comprehensive discussion and ablation study, and finally, Section VI concludes the paper.

## II. Related Work

In recent years, numerous methodologies have been proposed to address the challenges of small-object detection,

Published in the 34th International Conference on Electrical Engineering (ICEE 2026), pp. 430–434. The final published version is available on IEEE Xplore: https://doi.org/10.1109/ICEE70628.2026.11710697.

particularly in the UAV domain. These approaches can be broadly categorized into two primary streams: detection architectures and data augmentation techniques.

### A. UAV Detection Architectures

Prior research has predominantly concentrated on high-capacity models. For instance, Zhao et al. [8] assessed the performance of two-stage detectors, such as Faster R-CNN [7], using deep backbones like ResNet-50 [15], and demonstrated that network depth is crucial for extracting features from small targets. However, deploying such complex architectures on edge hardware, including Jetson or Raspberry Pi modules, faces significant computational constraints [9], [10]. These limitations necessitate adopting more optimized, lightweight models, such as the YOLOv11 family [11].

### B. Data Augmentation Strategies in UAV Imagery

Data augmentation techniques play a pivotal role in improving the performance of drone detection models, especially in data-constrained scenarios. A widely adopted approach in recent studies is synthetic sample generation. Laroca et al. [12] employed the Copy-Paste technique to extract drone instances from source images and superimpose them onto diverse backgrounds to mitigate data scarcity. In a separate study, Zheng et al. [13] evaluated two distinct approaches—Mask Copy-Paste and Rough Copy-Paste—to optimize this process for foggy environments. Their empirical findings revealed that the Rough method, despite its simplicity, achieves competitive accuracy (89.9%). It is important to note that these results were reported using the heavy YOLOv5L [16] model. Drawing upon this finding, this research utilizes Rough Copy-Paste as a baseline. However, the results of the present study indicate that while Copy-Paste is effective for high-capacity models, it induces synthetic artifacts and overfitting in lightweight architectures like the Nano model. Conversely, global transformation methods such as MixUp [17], which rely on pixel-level blending, may obscure critical details needed to detect tiny objects. Consequently, this research focuses on a Context-Aware approach based on the Mosaic technique [14], which, unlike Copy-Paste, preserves the semantic continuity of the image texture.

## III. Proposed Method

In this section, the proposed methodology for real-time UAV detection is elaborated. First, the rationale for selecting the base model architecture under hardware constraints is discussed. Subsequently, a critical analysis of object-based augmentation strategies is presented to highlight their limitations on lightweight models. Finally, the proposed Context-Aware Pipeline is introduced, which integrates image-level augmentation techniques to enhance model robustness. The overall framework of this approach is illustrated in Fig. 1.

### A. Model Architecture

Given the hardware constraints of edge processors, the network architecture was selected to balance accuracy and processing cost. In this study, the YOLOv11 Nano [11] model was chosen as the processing core. By leveraging optimized C3k2 blocks and the C2PSA attention mechanism introduced in this architecture [11], it extracts richer semantic features than previous versions, such as YOLOv8 [18], while requiring only a modest increase in parameters. However, the limited learning capacity of this model makes it vulnerable to input noise, a critical issue that served as the basis for designing the data augmentation strategy in this research.

### B. Analysis and Enhancement of Copy-Paste Strategy

In prior studies, Copy-Paste techniques have been recognized as a widely adopted strategy for increasing sample counts. In this study, the Rough Copy-Paste [13] strategy was used as the baseline. However, for better evaluation and to enhance sample quality, simple cropping and pasting were insufficient; therefore, an optimization module comprising photometric adjustments (Gamma, Brightness), random pixel dropout, and smart placement logic was added to the process, as illustrated in Fig. 1.

Despite applying these optimizations, experimental results demonstrated that the Copy-Paste technique creates structural challenges for lightweight models:

*1) High-Frequency Artifacts:* Pasting the target image onto the background, regardless of execution precision, leads to visual discontinuities and unnatural edges at the object boundary. Our experiments proved that the Nano model, due to its low capacity, tends to overfit on these artificial edges.

*2) Preprocessing Computational Overhead*: Generating high-quality samples (including loading, photometric changes, and logical placement) imposes a significant computational burden on the preprocessing stage.

### C. Proposed Strategy: Context Integrity

To overcome the significant challenges inherent in Copy-Paste methods, an alternative pipeline focusing on global transformations has been designed. As shown in Fig. 1, this strategy consists of two main components applied sequentially to the data during the training process:

*1) Mosaic Augmentation:* In this method, four training images are randomly selected and stitched into a 2×2 grid [14]. Beyond preserving the local context and preventing the creation of artificial edges around the target, this technique increases data diversity within each batch. In contrast to prior studies [12] that used Mosaic with a low probability (0.1) as a supplementary technique, this research adopts it with a significantly higher probability (0.75) as the core component, effectively replacing object-based methods.

*2) Photometric Adaptation (HSV):* At the end of the pipeline, controlled variations are explicitly applied to the Value (Brightness) component of the HSV color space, while Hue and Saturation remain unchanged to preserve semantic color information. This adjustment effectively simulates diverse illumination conditions (such as overexposure or shadows) without altering the object's structure. This adaptation is designed to enhance the model's stability against photometric inconsistencies, acting as a complementary mechanism to ensure generalization across varying environmental conditions.

## IV. Experiments And Results

In this section, the efficacy of the proposed method is evaluated against other strategies across various scenarios. First, the employed datasets and implementation details are described, followed by a quantitative and qualitative performance analysis.

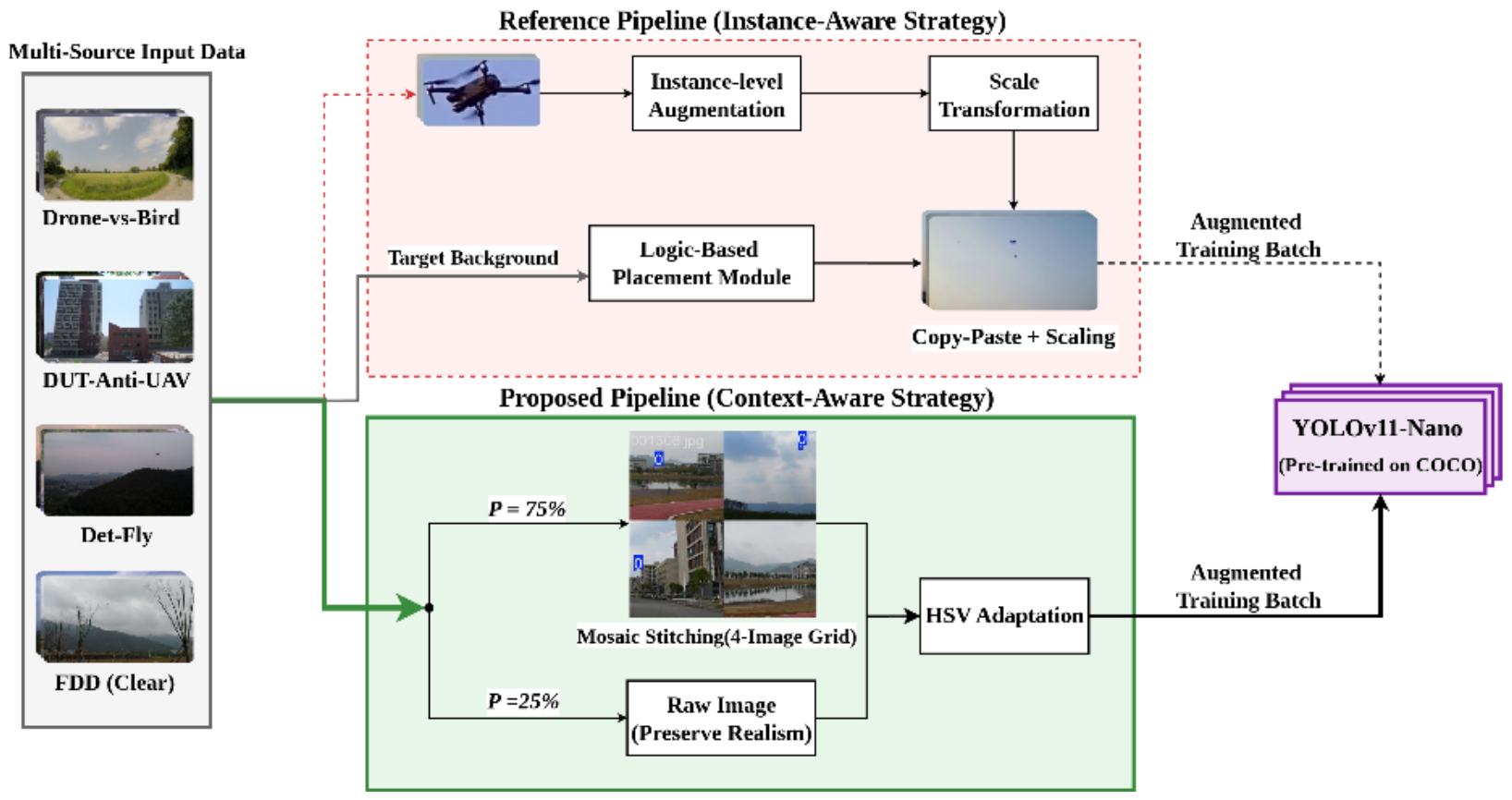

Fig. 1. Overview of the proposed framework. The upper section (Reference Pipeline) illustrates the instance-aware strategy based on Copy-Paste. The lower section (Proposed Pipeline) depicts the Context-Aware approach, which trains the YOLOv11 Nano model via probabilistic Mosaic (P=0.75) and Photometric Adaptation (HSV).

## *A. Datasets and Evaluation Metrics*

To ensure a comprehensive evaluation and establish the domain independence of the results, benchmark datasets specialized in small UAV detection with diverse characteristics have been utilized. A unifying challenge across all these datasets is the tiny scale of the targets, where the UAV often occupies a negligible portion of the image pixels (usually less than 1% of the frame), resulting in significant textural information loss. The details of each dataset and the evaluation metrics are described below:

*1) Drone-vs-Bird (DvB):* Introduced by Coluccia et al. [5], this dataset is considered one of the most challenging benchmarks for distinguishing drones from birds. It comprises 77 video sequences with varied backgrounds. In this study, to reduce information redundancy and optimize the training process, the frame sampling rate was set to 0.5 seconds, and the final evaluation was conducted on the exact 7 video sequences specified in [12].

*2) DUT-Anti-UAV:* Developed by Zhao et al. [8], this dataset contains over 10,000 images (split into train, val, and test sets) for detection. Its distinguishing feature is the high diversity of targets (including diverse drone models such as the DJI series) and the breadth of environmental conditions (day, night, cloudy, and complex urban backgrounds), which assesses the model's ability to identify various drone types under different lighting conditions.

*3) Det-Fly:* This dataset [4] comprises more than 13,000 high-resolution images from an Air-to-Air scenario. The images are characterized by diverse backgrounds categorized into four distinct subsets: Mountain, Urban, Field, and Sky. The significance of this dataset lies in simulating aerial engagement scenarios and severe changes in the camera's viewing angle relative to the target.

*4) Foggy Drone Dataset (FDD):* To assess model stability under adverse weather conditions, the FDD [13] was employed. This collection includes 7,187 synthetic foggy images generated using the Atmospheric Scattering Model. In this research, the Clear version was used for training, and the Foggy version was used exclusively for testing to evaluate the model's Domain Generalization capability.

*5) Evaluation Metrics:* To quantitatively evaluate model performance, two standard metrics, Precision and mean Average Precision (mAP@50), were used. The mAP@50 metric reflects the overall detection accuracy calculated at an Intersection over Union (IoU) threshold of 0.5, whereas Precision indicates the model's reliability in minimizing false positives, which is critical for security applications.

## *B. Implementation Details*

All experiments utilized the YOLOv11 Nano [11] model, trained on an NVIDIA RTX 5070 Ti (16GB) GPU. While high-end hardware was used for training, the Nano architecture ensures it is feasible for edge deployment. Training started with COCO [19] pre-trained weights. To minimize the distribution shift between augmented and real data, Mosaic augmentation was turned off for the final 10 epochs to allow fine-tuning on natural images. An Early Stopping mechanism was also applied. Comprehensive hyperparameter settings are detailed in Table I.

Regarding HSV augmentation, Hue and Saturation gains were turned off (h=0, s=0) to preserve the intrinsic color. However, the Value gain was adaptively set to [0,0.4] based on domain characteristics: a minimal gain ($v \approx 0$) was applied to the high-luminance FDD dataset to prevent saturation, while a higher gain (v=0.4) was used in clear-weather domains to maximize illumination robustness. For reproducibility, the source code and configuration files are publicly available[1].

TABLE I
HYPERPARAMETERS USED IN THE TRAINING PROCESS.

| Parameter | Value | Parameter | Value |
|---|---|---|---|
| Model | YOLOv11n | Optimizer | SGD |
| Input Size | 640x640 | Momentum | 0.9 |
| Batch Size | 32 | Weight Decay | 0.0005 |
| Epochs | 100 | Base LR | 0.01 |
| Patience | 15 | Close Mosaic | 10 |

## *C. Comparative Evaluation and Performance Analysis*

For rigorous validation, the performance of the proposed method was compared with five strategies prevalent in state-of-the-art studies, detailed in Table II: 1) Baseline: Training without augmentation; 2) Heavy Instance-Aug: Copy-Paste with aggressive geometric and color transformations [12]; 3) Pixel-Level Aug: Global geometric and photometric changes [12]; 4) Instance-Aware: Rough Copy-Paste with Scale Jittering [13]; and 5) Global Mixing: MixUp [17] (P=0.1) to evaluate image blending while preserving small target visibility.

[1] https://github.com/amirzamanii/Context-Aware-UAV-Detection

TABLE II
QUANTITATIVE COMPARISON OF DATA AUGMENTATION STRATEGIES ACROSS FOUR SMALL UAV DETECTION BENCHMARKS. VALUES REPRESENT MEAN AVERAGE PRECISION (mAP@50), WITH THE BEST RESULTS MARKED IN BOLD.

| Method (Strategy) | Augmentation Details | DvB | DUT | Det-Fly | FDD | Param (M) | GFLOPs |
|---|---|---|---|---|---|---|---|
| Baseline | None | 50.6 | 79.9 | 82.6 | 40.5 | ~2.6 | 6.3 |
| Heavy Instance-Aug [12] | Copy-Paste + Global Distortions | 61.8 | 85.2 | 84.2 | 52.9 | ~2.6 | 6.3 |
| Pixel-Level Aug [12] | Spatial & Photometric Transforms | 63.7 | 85.8 | 86.3 | 52.6 | ~2.6 | 6.3 |
| Instance-Aware [13] | Copy-Paste + Scale Jittering | 52.5 | 78.3 | 84.1 | 44.5 | ~2.6 | 6.3 |
| Global Mixing | MixUp (P=0.1) | 47.8 | 78.7 | 81.9 | 52.7 | ~2.6 | 6.3 |
| **Context-Aware (Ours)** | **Mosaic + HSV Adaptation** | **68.0** | **86.9** | **90.9** | **58.7** | ~2.6 | 6.3 |

Furthermore, the analysis of Table II demonstrates the decisive superiority of the proposed Context-Aware method across four benchmark datasets without any computational trade-off. As shown, all methods maintain a constant parameter count (~2.6M) and computational complexity (6.3 GFLOPs) because the augmentations are applied strictly offline. A primary finding is the structural mismatch between heavy augmentations and lightweight models; for instance, in the challenging DvB dataset, Copy-Paste performed significantly worse (52.5%) than our method (68.0%) due to synthetic artifacts. In contrast, by preserving semantic consistency, our approach achieved high performance stability in diverse environments, from complex urban DUT scenes to mountainous Det-Fly landscapes.

## V. Discussion

In this section, the experimental findings are interpreted to provide deeper insights into the underlying mechanisms of the proposed method compared to state-of-the-art strategies. The analysis focuses on the contributions of individual components, the structural reasons for the failure of instance-level augmentations, and robustness behavior under environmental degradation.

### A. Ablation Study

The ablation study in this section focuses exclusively on the DvB dataset, which serves as the most challenging benchmark for small-object detection. To quantify the specific contribution of each component, the results reveal a clear hierarchy:

*1) Dominant Role of Mosaic:* Integrating Mosaic augmentation alone yielded a substantial performance boost, elevating mAP from 50.6% (Baseline) to 65.2%. This result confirms that the primary capabilities of the proposed pipeline—learning scale-invariant features and preserving local context—are driven by the Mosaic strategy, which effectively addresses the small-object problem without introducing artifacts.

*2) Complementary Role of HSV:* The addition of HSV adaptation acted as an auxiliary regularizer, raising the final mAP to 68.0%. This synergy demonstrates that while Mosaic addresses structural challenges (scale and edges), HSV ensures robustness against photometric variations, and their combination is essential for optimal performance.

*3) Synergy:* The full pipeline achieves optimal performance, demonstrating that Mosaic provides the structural generalization while HSV ensures photometric invariance, creating a balanced solution for diverse environments.

### B. Robustness Analysis in Adverse Weather

The evaluation on the FDD dataset (Table II and Fig. 2) revealed critical insights into model generalization:

*1) Precision vs. mAP in Copy-Paste:* As shown in Fig. 2, the Copy-Paste method achieves the highest Precision (80.9%) on the FDD dataset. However, its overall performance in Table II shows a notably low mAP (44.5%). This discrepancy occurs because Copy-Paste relies heavily on sharp, artificial edges. In foggy conditions, the model becomes overly conservative; it only detects highly visible targets—resulting in high Precision—but fails to detect obscured drones. This severe drop in Recall drastically reduces its mAP compared to other methods.

*2) Behavior of MixUp:* As illustrated in Fig. 2, the MixUp strategy exhibits a distinct behavior. While its Precision (79.2%) is marginally lower than the Copy-Paste method (80.9%), it achieves a significantly higher mAP (52.7% vs. 44.5%). This performance disparity indicates that MixUp, by creating visual ambiguity akin to fog, forces the model to be cautious and suppresses false alarms, thereby avoiding the severe miss rate (low recall) observed in Copy-Paste methods.

*3) Visual Verification:* Qualitative results in Fig. 3 depict model robustness. Row (a) shows that the Copy-Paste method struggles with edge degradation, detecting targets with noticeably lower confidence. In contrast, Row (b) demonstrates the proposed method's ability to accurately localize drones with robust confidence by leveraging rich contextual information, proving its operational reliability.

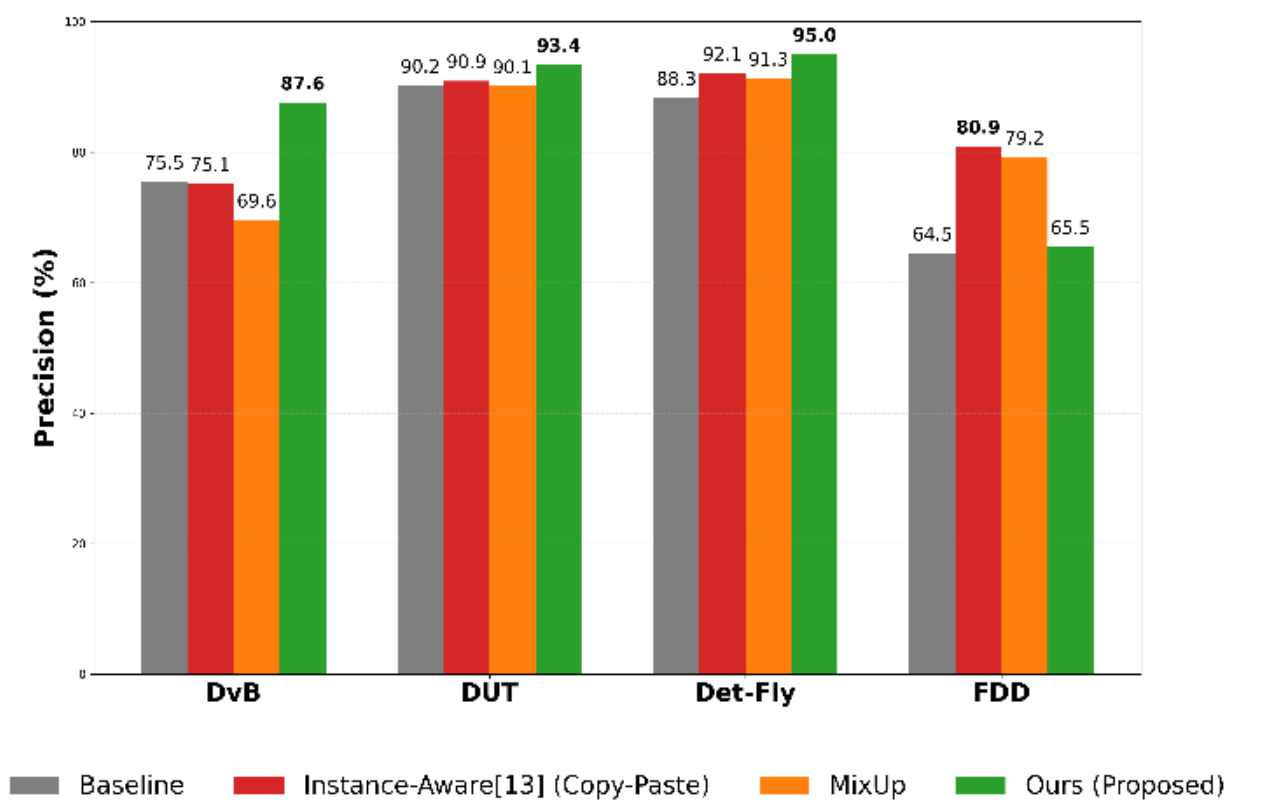

Fig. 2. Precision comparison across four datasets, highlighting the trade-off with mAP@50 results in Table II.

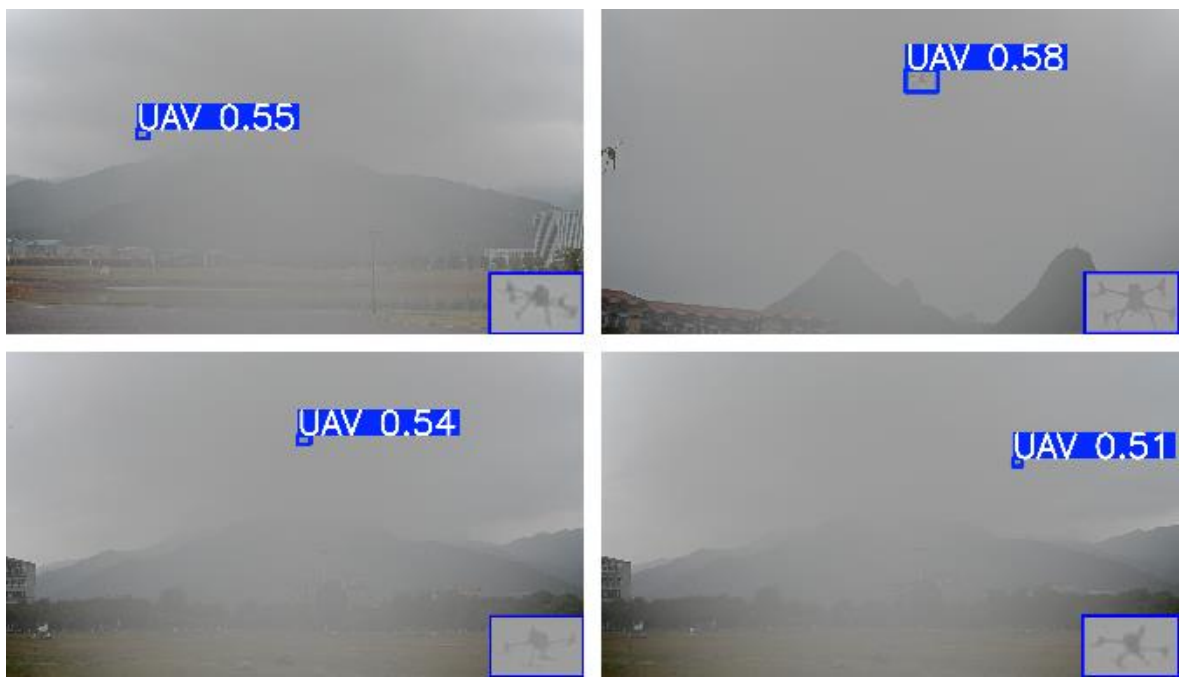

(a)

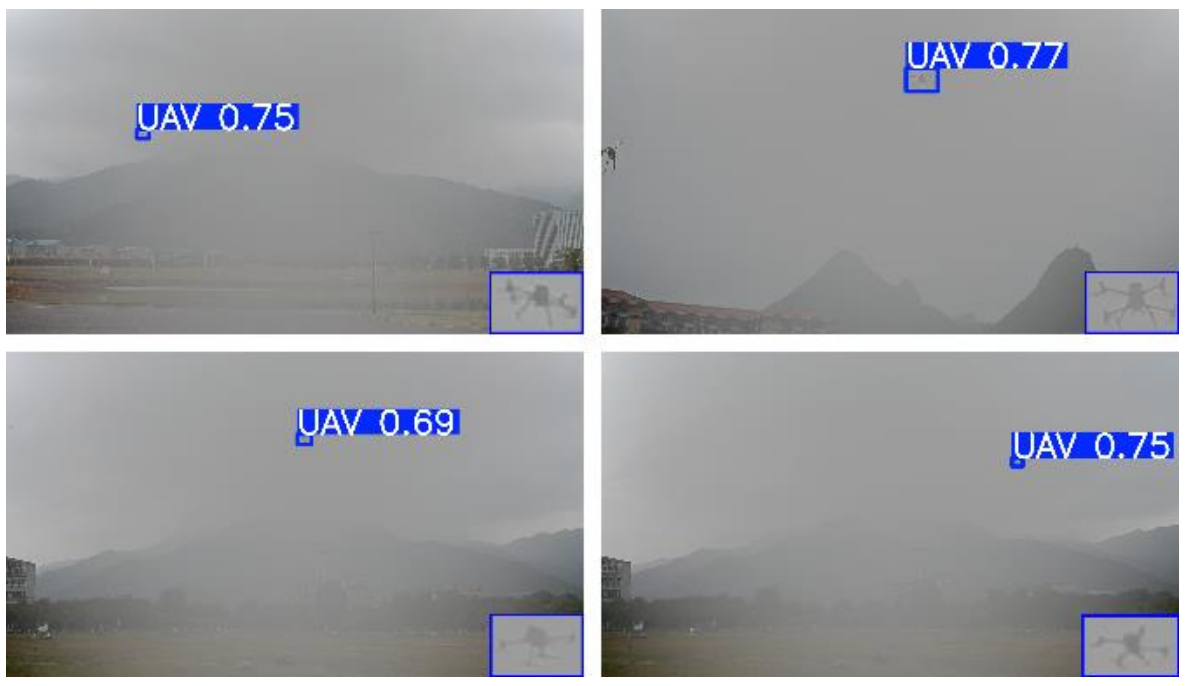

(b)

Fig. 3. Qualitative comparison of the Copy-Paste method versus the proposed approach on the FDD dataset. (a) Copy-Paste results. (b) Proposed method results on corresponding frames. Zoomed insets highlight targets; bounding boxes indicate predicted locations and confidence scores.

## VI. Conclusion

The primary objective of this research was to achieve accurate, real-time small UAV detection on resource-constrained hardware. Empirical results demonstrated that heavy data augmentation strategies, such as Copy-Paste, while effective for large models, degrade performance in lightweight architectures (e.g., YOLOv11n) by introducing noise and artificial edges. In contrast, the proposed context-aware approach (combining Mosaic and HSV adjustments) significantly enhanced detection accuracy without imposing any additional computational overhead. This ensures an optimal speed-accuracy balance for edge systems. Furthermore, investigations revealed that under specific conditions, such as fog, the MixUp strategy serves as an effective complement to suppress false alarms. Future work will focus on expanding augmentation methods to overcome the performance ceiling of standalone strategies. Furthermore, we aim to evaluate real-time deployment on hardware platforms like NVIDIA Jetson using TensorRT and Raspberry Pi 5 via NCNN to ensure maximum efficiency across diverse edge environments.

## AI Usage Statement

The authors declare that they used artificial intelligence tools solely to improve and edit the article's text.